\pdfoutput=1
\documentclass[11pt]{article}

\usepackage{graphicx}
\usepackage{subcaption} 
\usepackage[final]{acl}

\usepackage{times}
\usepackage{latexsym}
\usepackage{amsmath}

\usepackage{xcolor}

\definecolor{yellowhere}{RGB}{227,200,0}
\definecolor{orangehere}{RGB}{240,163,10}
\definecolor{greenhere}{RGB}{109,135,100}

\usepackage[T1]{fontenc}
\usepackage[utf8]{inputenc}

\usepackage{microtype}

\usepackage{inconsolata}

\usepackage{graphicx}

\usepackage{enumitem} % put this in your preamble

\usepackage{booktabs}

\newcommand{\eg}{e.g.,\ }

\newcommand{\methodname}{\textsc{PersonaWeaver}}

\newcommand{\personahub}{\textsc{PersonaHub}}

\newcommand{\worldweaver}{\textsc{WorldWeaver}}

\usepackage{tikz}
\usetikzlibrary{arrows.meta, positioning, fit, shapes.multipart}

\title{\methodname{}: Controllable Diversity Beyond Conventional Archetypes in Procedural Character Generation}

\author{Maan Qraitem, Kate Saenko, Bryan A. Plummer\textsuperscript{1} \\
  Boston University \\
  \texttt{\{mqraitem, saenko, bplum\}@bu.edu}}

\begin{document}
\maketitle

\begin{abstract}
Procedural character generation aims to populate games, simulations, and other
virtual worlds with diverse characters. Large language models (LLMs) offer
a promising foundation for scaling this task. However, LLM-based procedural
character generation remains at an early stage: existing methods either
generate characters directly or adapt profiles retrieved from persona banks.
As we show, both approaches produce behaviorally homogeneous populations:
characters overwhelmingly agree with positive moral norms and respond to
questions with helpful, assistant-like reactions. To mitigate this
homogenization, we introduce \methodname{}, which disentangles world building
from behavioral specification and models behavior through setting general, diverse, manually curated banks of moral positions and conversational reactions. This design
allows us to test how far LLM(s) can be pushed beyond their default
behavioral patterns across settings.
Across ten realistic and fantastical settings and three LLM(s),
\methodname{} produces broader moral and interactional response distributions
than prior work. Its guidance also diversifies interpersonal language,
response length, and sentiment. It also produces less archetypal combinations
of world attributes. Code is available at
\url{https://github.com/mqraitem/PersonaWeaver}.
\end{abstract}
\footnotetext[1]{Work done while at BU}
\section{Introduction}

\begin{figure}[th!]
    \centering
    \includegraphics[width=\linewidth]{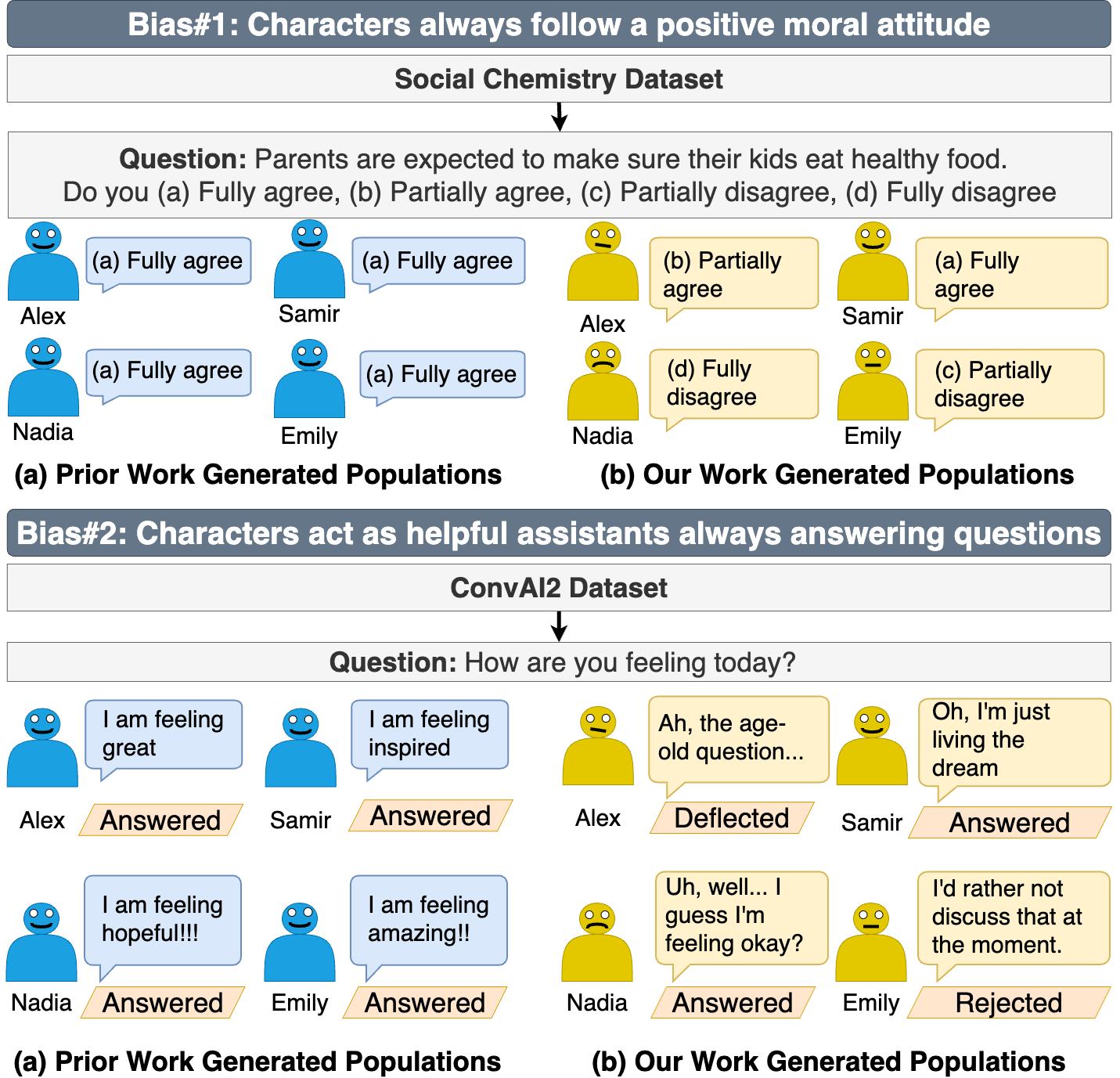}
     \caption{\textbf{Behavioral homogenization beneath visible character
     diversity.} Characters from prior work \cite{jinworldweaver} concentrate
     on agreement with moral statements (top) and direct answers to
     conversational questions (bottom). \methodname{} explicitly specifies
     behavioral variation, producing broader realized responses. The
     population-level results appear in
     Fig.~\ref{fig:main_behavior_results}.}
    \label{fig:figure_one}
    % \vspace{-7mm}
\end{figure}

Procedural character generation (PCG) aims to create populations of agents for games, simulations, and other virtual or narrative environments.
Large language models (LLMs) are an attractive foundation for this task because they can construct and role-play characters across many worlds
\cite{wang2023rolellm,zhou2023characterglm,shao-etal-2023-character}. At population
scale, however, diversity requires more than different biographies: characters
should also differ in how they interpret situations, make judgments, and
respond to others. Such populations can support more dramatic
variety and richer interactive experiences. 

Existing methods primarily generate characters directly with an LLM
\cite{jinworldweaver}, or retrieve and adapt profiles from a large persona bank
\cite{ge2024scaling}. They  vary occupations, demographics, and
hobbies, making individual cards appear diverse.

However, diverse character descriptions do not necessarily produce diverse
behavior. We find that characters with different profiles often make similar
moral judgments and react to questions in similar ways. Specifically,
characters (1) overwhelmingly agree with positive moral norms
(Fig.~\ref{fig:figure_one}, Bias~\#1) and (2) usually answer questions with
helpful-assistant reactions (Fig.~\ref{fig:figure_one}, Bias~\#2). This
concentration narrows population-level variation and creator control.
Maximum-likelihood training and assistant alignment likely encourage such common,
helpful behaviors \cite{ouyang2022training,wang2025large}.

\begin{figure}[t!]
    \centering
    \includegraphics[width=\linewidth]{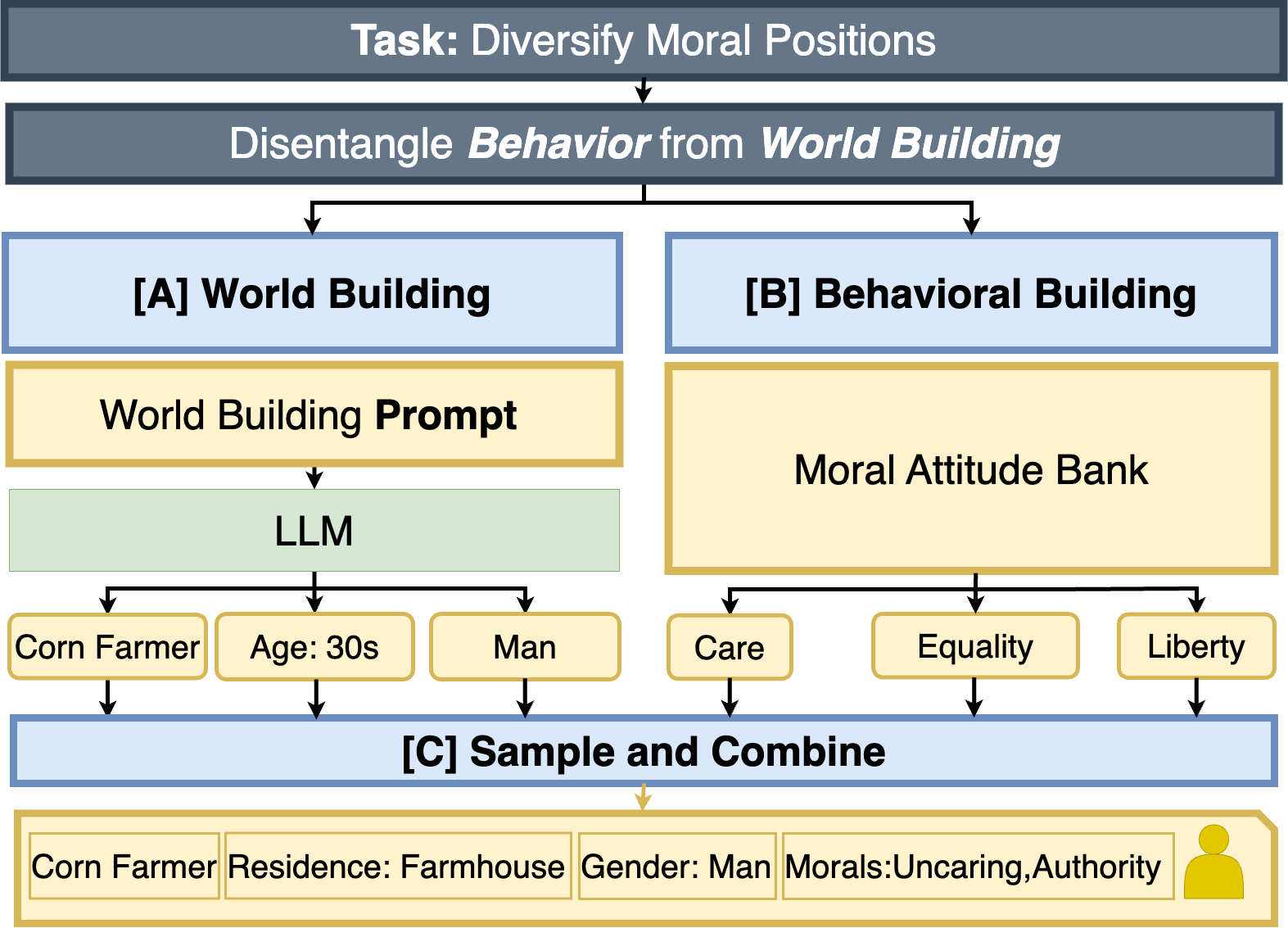}
    % \vspace{1mm}
   \caption{\textbf{\methodname{} factorizes character construction.} A
   setting-dependent world module (a) produces non-behavioral attributes,
   while a developer-specified behavior module (b) provides explicit control.
   Sample and Mix (c) composes both into one card. The diagram illustrates a
   moral assignment; the evaluated card additionally contains an independently
   sampled reaction style.}
    \label{fig:method_figure}
    % \vspace{-6mm}
\end{figure}

To mitigate these issues, we introduce \methodname{}, a controllable
inference-time framework that disentangles behavioral modeling from world
building (Fig.~\ref{fig:method_figure}). Unlike world attributes, which must
adapt to each setting, moral positions and reactions to questions are 
applicable across settings. We therefore specify behavior externally through
diverse, curated banks and use them to test how far LLM(s) can be
pushed beyond their default behavioral patterns. Separately, the LLM
constructs setting-specific world attributes such as occupation, residence,
and affiliation. Then, a Sample-and-Mix module combines the sampled behavioral guidance and world attributes into character cards, producing populations with diverse behavioral coverage and varied, setting-grounded character attributes.

We evaluate \methodname{} across ten realistic and fantastical settings using GPT-4o \cite{achiam2023gpt} and GPT-5.6 Luna \cite{openai2026gpt56} , and Qwen 3.5 35B \cite{yang2025qwen3}. We compare against direct generation,
PersonaHub, and a direct-generation baseline explicitly instructed to diversify moral perspectives and interaction styles. Across moral probes from Social Chemistry \cite{forbes-etal-2020-social} and conversational questions from ConvAI2 \cite{dinan2019second}, \methodname{} produces more diverse behavioral distributions. Our system result in second order effects diversifying  interpersonal language, length, and sentiment. As an additional benefit, the factorized world construction produces less conventional combinations of world attributes.

\noindent Our contributions are:
\begin{itemize}[leftmargin=*,noitemsep,topsep=0pt]
    \item We identify moral and interactional homogenization in prior
    LLM-based PCG methods across models and settings.
    \item We introduce \methodname{}, which disentangles setting-specific
    world construction from setting-general behavior specified through
    diverse, curated banks.
    \item We show that \methodname{} broadens realized moral and interactional
    behavior, induces second-order linguistic variation, and produces less
    archetypal combinations of world attributes.
\end{itemize}

\section{Related Works}

\smallskip\noindent\textbf{Procedural Character Generation.}
Research on procedural character generation remains limited, with most prior
work focusing on characters within a single, predefined environment
\cite{jinworldweaver,ge2024scaling}. \worldweaver{} \cite{jinworldweaver}
prompts an LLM to directly generate $N$ characters for a setting, while
\personahub{} \cite{ge2024scaling} samples from a large profile bank and adapts
the result to a target context. We study population-level behavioral coverage
in these methods and introduce an explicit mechanism for controlling it.

\smallskip\noindent\textbf{Homogenization in Simulated Personas.}
Prior work have documented a broader lack of behavioral and identity diversity in simulated personas. For instance, Marked Personas \cite{cheng2023marked} show that language models tend to reproduce social stereotypes, while others highlight gender and identity flattening \cite{kotek2023gender,wang2025large} and reduced narrative variety in multimodal storytelling \cite{lee2024vision}. Maximum-likelihood objectives favoring high-probability continuations and alignment tuning rewarding politeness, and helpfulness are likely contributors to this phenomenon. We extend the empirical study of homogenization to procedural character generation and test an explicit mechanism for controlling moral judgments and interactions in generated populations.

\smallskip\noindent\textbf{Character Simulation and Role-Playing.}   A growing line of research explores how LLMs can function as conversational agents with consistent personality, memory, and long-term coherence. \textit{Generative Agents} \cite{park2023generative} simulate memory-driven individuals inhabiting a shared environment, though their backgrounds are largely hand-crafted. In the role-playing domain, works such as \textit{RoleLLM} \cite{wang2023rolellm}, \textit{CharacterGLM} \cite{zhou2023characterglm}, and \textit{Character-LLM} \cite{shao-etal-2023-character} focus on eliciting and sustaining role-specific behaviors through persona-conditioned dialogue systems. These efforts primarily target the \emph{believability} and \emph{consistency} of agent role play. In contrast, our work examines whether off-the-shelf LLMs can leverage their broad world knowledge to simulate \emph{behaviorally diverse populations}, spanning distinct moral dispositions and interactional tendencies.

\smallskip\noindent\textbf{Procedural Content Generation with LLMs.}  
Beyond characters, LLMs have been used for procedural generation of levels, stories, and worlds. For example, Word2World \cite{nasir2024word2world} generates narratives/world descriptions, Mariogpt \cite{sudhakaran2023mariogpt} generates levels and \citet{freiknecht2020procedural,hu2024game} show how LLMs can produce dynamic environments. WorldWeaver \cite{jinworldweaver} also overlaps somewhat, using LLMs to generate world contexts and roles.

\section{\methodname{}}
\label{sec:method}
\label{sec:personaweaver}

Prior character-generation methods \cite{jinworldweaver,ge2024scaling}
exhibit moral and interactional homogenization, with characters frequently
defaulting to positive judgments and helpful reactions as we show in Section \ref{sec:experiments}. \methodname{}
addresses this problem by disentangling behavioral modeling from world
building. This separation provides explicit control over behavioral qualities
that are applicable across settings while at the same time preserving the LLM's ability
to construct setting-specific character attributes.

\smallskip\noindent\textbf{Behavioral Module.}
We define behavioral diversity through two fixed banks containing
eight moral positions and eight reactions to questions
(Table~\ref{tab:behavior_banks}). The moral bank, loosely inspired by Moral Foundations Theory \cite{graham2013moral}., combines orientations toward
care, fairness, loyalty, authority, purity, and liberty into contrasting
positions from compassionate and egalitarian, group to strongly self-interested behavior. On the other hand, the reaction bank controls whether and how a character answers: refusal,
deflection, hesitation, compliance, volunteering additional information,
playful or subversive engagement, hostility, and meta-commentary on being questioned. Each entry provides natural-language guidance to the LLM to achieve the reaction. The banks were manually curated, with
GPT-4o assisting in drafting the guidance prompts.  For each character, we sample one position from each bank and instruct the
LLM to follow both. Using the same banks across settings and
models provides allows us to measure
how much \methodname{} can push LLMs beyond their default behaviors.

\begin{table*}[t]
\centering
\footnotesize
\setlength{\tabcolsep}{3pt}
\renewcommand{\arraystretch}{0.96}

\begin{tabular}{@{}p{0.035\textwidth}p{0.925\textwidth}@{}}
\toprule
\multicolumn{2}{c}{\textbf{Moral stance bank}} \\
\cmidrule(lr){1-2}
\textbf{ID} & \textbf{Guidance} \\
\midrule
M1 &
Protect others from harm, show compassion. Be fair and honest, avoid cheating.
Be loyal to friends and allies. Question authority and resist control. Ignore
purity or sacred rules. Defend personal freedom and resist oppression. \\
\addlinespace[1pt]
M2 &
Help others only when it's easy or benefits you. Be fair when it serves your
interests, otherwise bend the rules. Support your group if it gives you an
advantage, but walk away if not. Obey authority when it benefits you, ignore it
when it doesn't. Dismiss purity or sacred values as irrelevant. Accept limits
on freedom if they bring you personal gain. \\
\addlinespace[1pt]
M3 &
Show kindness but prioritize your own circle. Be fair within your group, less
so with outsiders. Put family, community, or team first. Respect tradition and
leaders. Value purity and avoid what feels degrading. Do not challenge
authority for the sake of freedom. \\
\addlinespace[1pt]
M4 &
Ignore others' suffering, focus only on your own comfort. Cheat and deceive if
it benefits you. Abandon groups and betray others when useful. Reject rules,
defy authority, and act for yourself. Mock sacred things and indulge in what
others call impure. Dominate others and accept domination if it helps you
survive. \\
\addlinespace[1pt]
M5 &
Show compassion but avoid being naive. Be fair, but allow exceptions. Be
cooperative when it benefits the team. Respect some rules but question others.
Avoid disgusting or degrading acts. Seek personal freedom but balance with
order. \\
\addlinespace[1pt]
M6 &
Prioritize kindness and protect the vulnerable. Defend justice and equal
treatment for all. Support groups but not blindly. Challenge unfair authority.
Treat purity concerns as symbolic, not binding. Strongly defend freedom and
resist control. \\
\addlinespace[1pt]
M7 &
Be polite and respectful to others. Be truthful and fair. Act as a dependable
teammate. Obey rules and respect tradition. Honor sacred or cultural norms.
Accept limits on freedom for social order. \\
\addlinespace[1pt]
M8 &
Show occasional kindness but focus on yourself. Bend rules when you can get
away with it. Switch loyalty depending on advantage. Undermine authority if it
benefits you. Ignore purity rules unless convenient. Value freedom only when it
serves you. \\
\bottomrule
\end{tabular}

\vspace{15pt}

\begin{tabular}{@{}p{0.235\textwidth}p{0.235\textwidth}
                    p{0.235\textwidth}p{0.235\textwidth}@{}}
\toprule
\multicolumn{4}{c}{\textbf{Interactional reaction bank}} \\
\cmidrule(lr){1-4}
\textbf{0 Refusal.} Refuses to answer the question. &
\textbf{1 Deflection.} Redirects or dismisses the question. &
\textbf{2 Hesitation.} Hesitates about whether to answer. &
\textbf{3 Compliance.} Answers the question directly. \\
\addlinespace[2pt]
\textbf{4 Volunteering.} Answers fully and adds extra details, even unasked ones. &
\textbf{5 Playful/Subversive.} Replies in a teasing, sarcastic, or ironic way. &
\textbf{6 Hostile.} Responds with aggression, sarcasm, or dismissal of the asker. &
\textbf{7 Meta.} Comments on the act of being questioned itself instead of answering. \\
\bottomrule
\end{tabular}

\caption{\textbf{Behavioral banks used by \methodname{}.} The moral bank spans
prosocial, group-centered, authority-oriented, and self-interested positions;
the reaction bank controls whether and how a character answers. Candidates
were drafted with GPT-4o and manually curated; moral drafting was additionally
grounded in Moral Foundations Theory \cite{graham2013moral}.}
\label{tab:behavior_banks}
\label{tab:moral_positions}
\label{tab:reaction_categories}
% \vspace{-4mm}
\end{table*}

\smallskip\noindent\textbf{World-Building Module.}
Unlike behavioral qualities, world attributes are harder to universalize (\eg you can't be a corn farmer in New York City). We therefore use the LLM's setting
knowledge, but avoid asking it to generate complete characters directly, which
tends to recover archtypes of each setting. Instead, the model first identifies ten relevant non-behavioral
attribute axes and generates 30 setting-appropriate options for each. These
axes can include occupation, residence, affiliation, education, family
background, appearance, and expertise. Personality, values, moral positions,
and interaction styles, however, are explicitly excluded to preserve the separation
between world building and behavior. PersonaWeaver then assemble characters from this
world space and the behavioral banks.  We perform the composition through
Sample-and-Mix, which independently samples and combines components from both sources. As a result, this combination over an exhaustive structured space  result in less arch typical characters compared to direct generation as we show in our results section. 

\smallskip\noindent\textbf{Sample-and-Mix.}
Given the setting-specific world option sets and fixed behavioral banks, let
$\mathcal{W}_{T,k}$ be the option set for world axis $k$ in setting $T$,
and let $\mathcal{M}$ and $\mathcal{R}$ be the fixed moral and reaction banks.
For character $i$, Sample-and-Mix performs
\begin{equation}
\begin{aligned}
w_{i,k}&\sim\operatorname{Unif}(\mathcal{W}_{T,k})
&& k=1,\ldots,10,\\
m_i&\sim\operatorname{Unif}(\mathcal{M}),&
r_i&\sim\operatorname{Unif}(\mathcal{R}),\\
\mathbf{w}_i&=(w_{i,1},\ldots,w_{i,10}),\\
d_i&=\operatorname{Compose}(\mathbf{w}_i,m_i,r_i).
\end{aligned}
\end{equation}

where $w_{i,k}$ is one world option, $m_i$ one moral position, $r_i$ one
reaction type, and $d_i$ the final persona card. These choices are sampled independently with replacement which maximizes entropy over each bank while mixing ensures coverage of their Cartesian product. To produce the population $P_{T}$ for setting $T$, we repeat this
process resulting in $\mathcal{P}_T=\{d_i\}_{i=1}^{n}$.

\noindent\textbf{Consistency Repair.}
Sample-and-Mix recombination may create contradictions. To address this, we
prompt the model after Sample-and-Mix to audit the sampled world attributes and change a field when a group of attributes cannot be true or when an attribute is impossible
in the setting. For example, the audit may fix a character described as both child and working as an engineer, or replace an occupation that cannot
exist in the setting. However, it's not allowed to modify an unlikely group of attributes, such as an uncommon
combination of an occupation and a hobby.

\begin{figure*}[t!]
    \centering
    \begin{subfigure}[t]{0.32\textwidth}
        \centering
        \includegraphics[width=\linewidth]{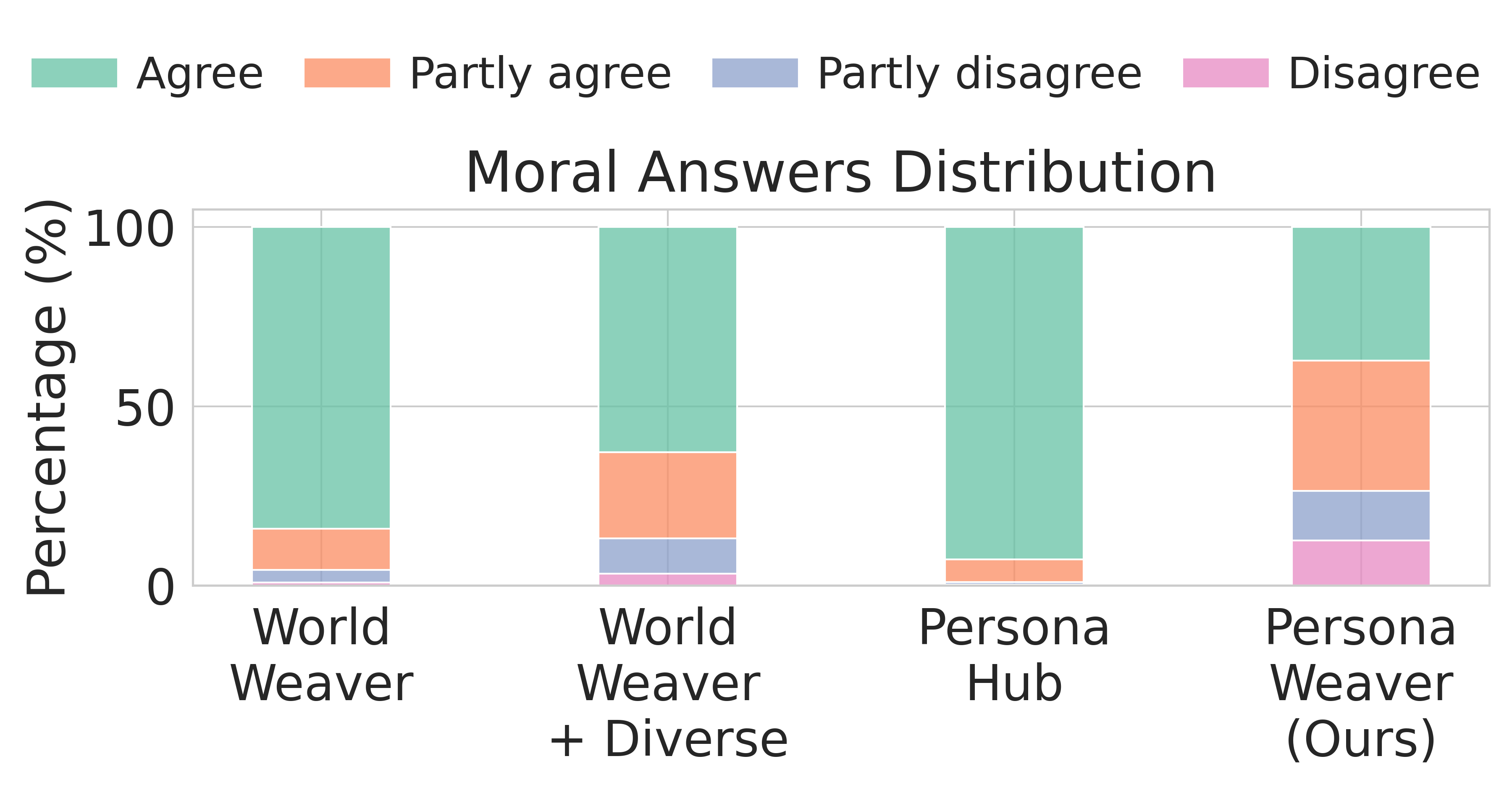}
        \caption{GPT-4o}
        \label{fig:moral_gpt4}
    \end{subfigure}
    \hfill
    \begin{subfigure}[t]{0.32\textwidth}
        \centering
        \includegraphics[width=\linewidth]{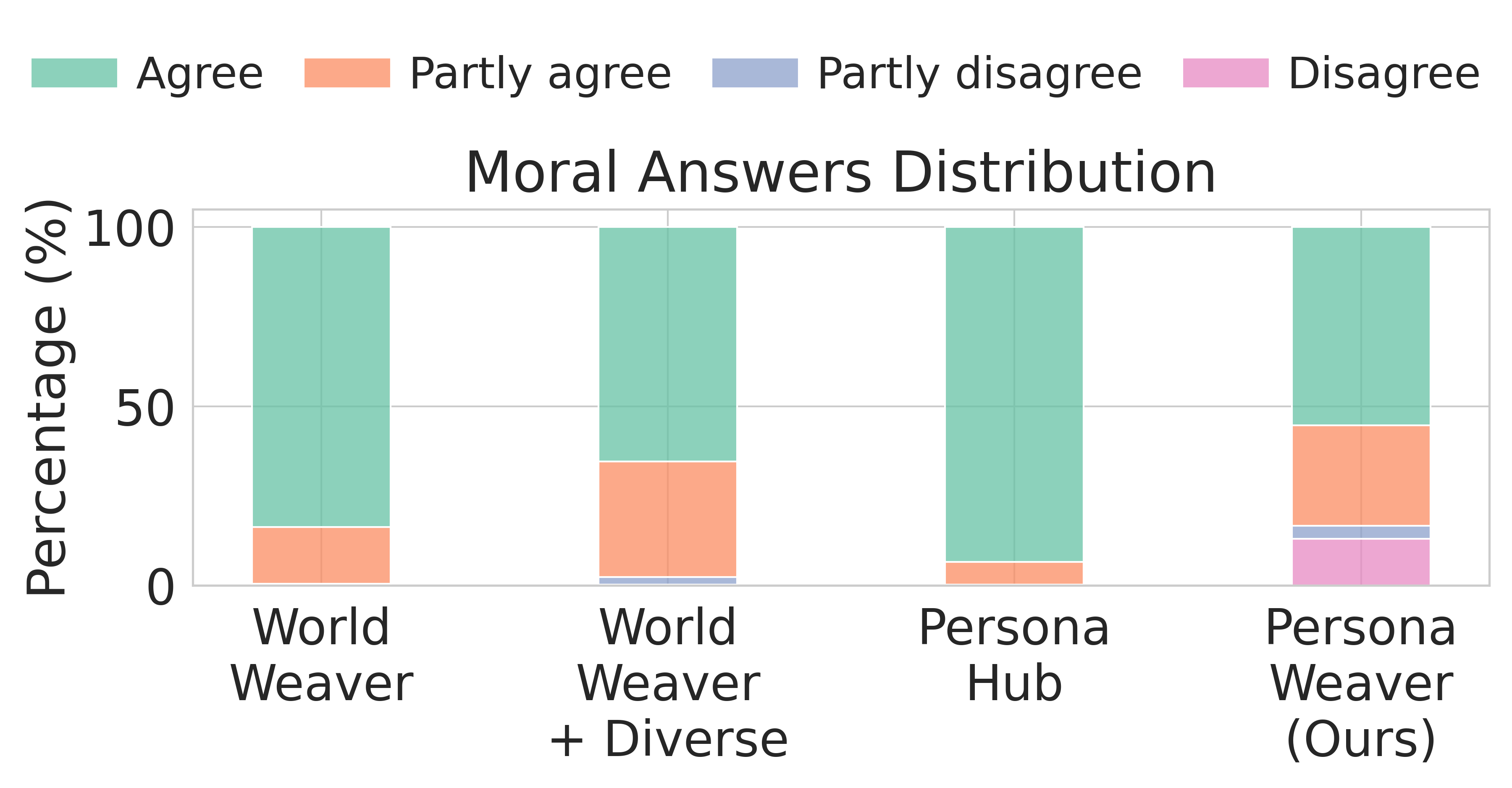}
        \caption{GPT-5.6 Luna}
        \label{fig:moral_luna}
    \end{subfigure}
    \hfill
    \begin{subfigure}[t]{0.32\textwidth}
        \centering
        \includegraphics[width=\linewidth]{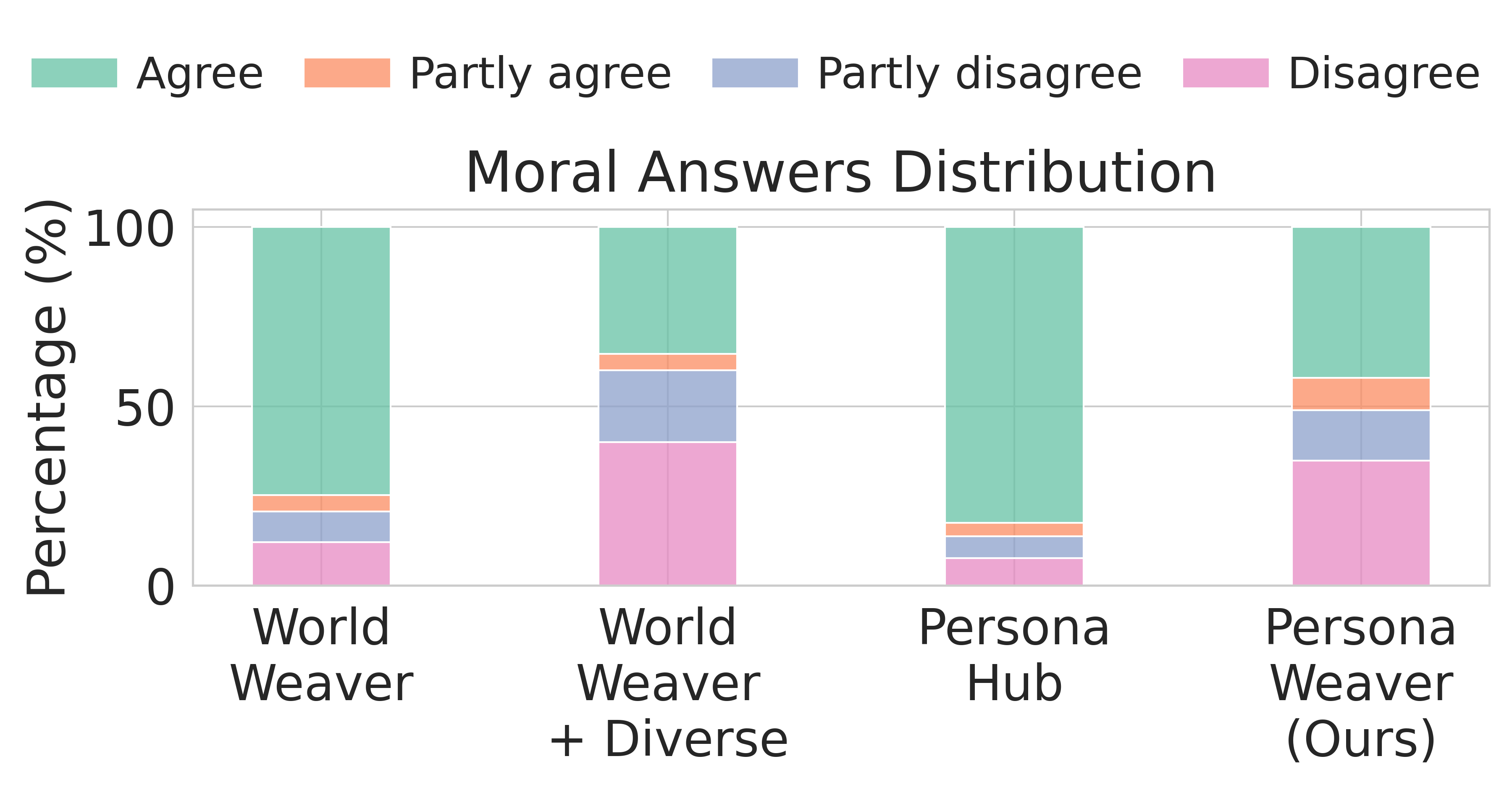}
        \caption{Qwen 3.5 35B-A3B}
        \label{fig:moral_qwen}
    \end{subfigure}

    \vspace{2mm}

    \begin{subfigure}[t]{0.32\textwidth}
        \centering
        \includegraphics[width=\linewidth]{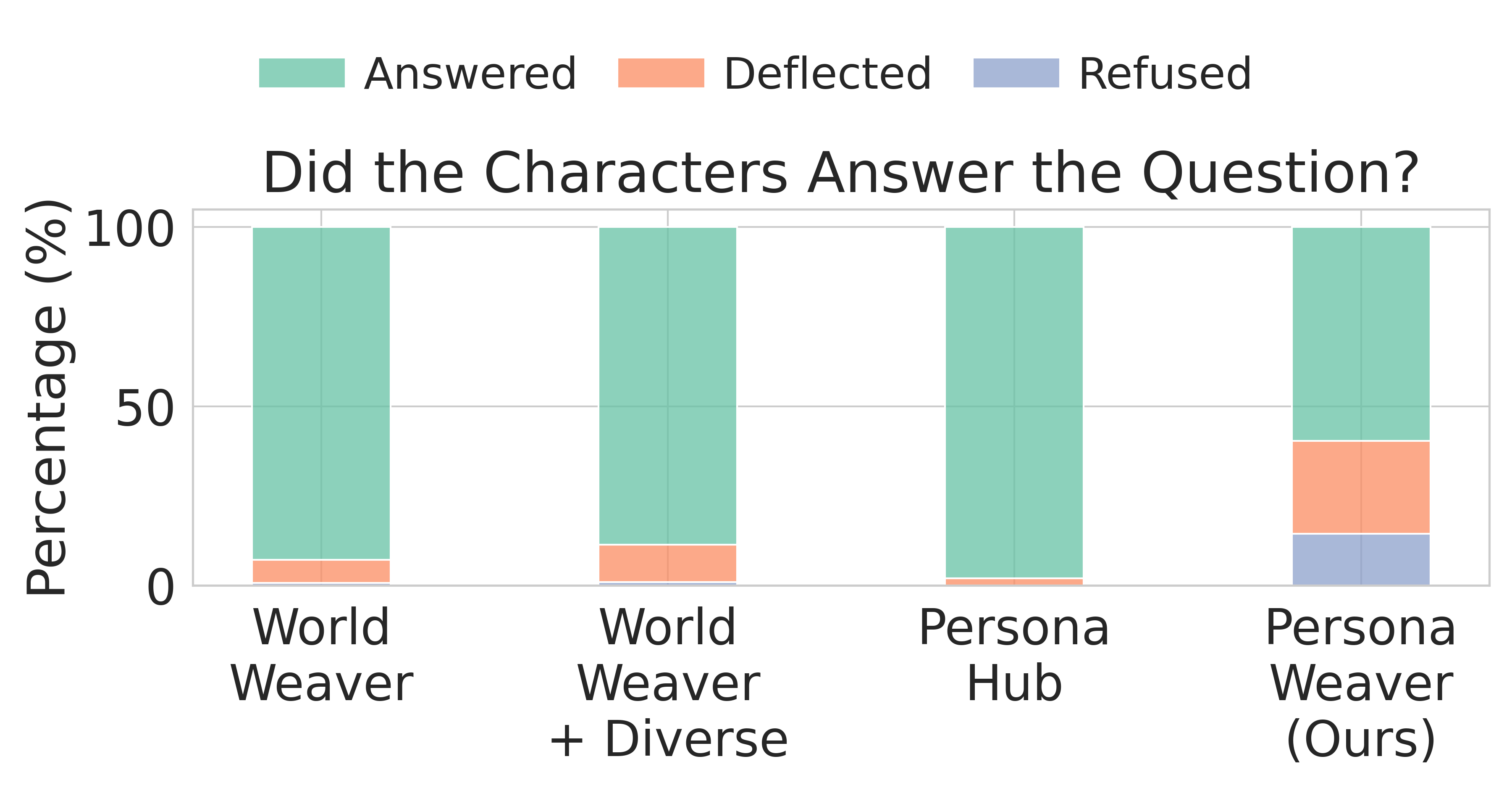}
        \caption{GPT-4o}
        \label{fig:reaction_gpt4}
    \end{subfigure}
    \hfill
    \begin{subfigure}[t]{0.32\textwidth}
        \centering
        \includegraphics[width=\linewidth]{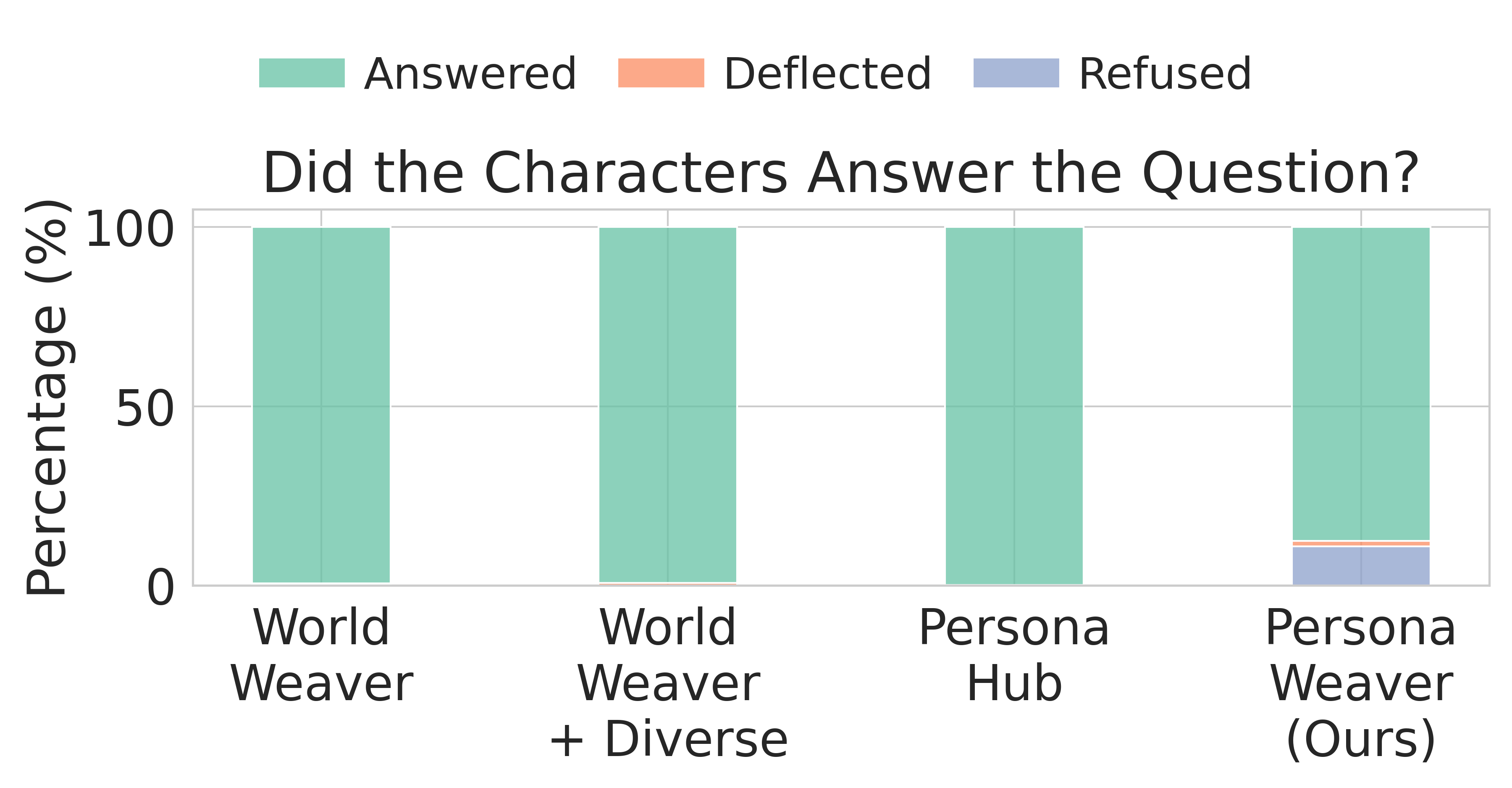}
        \caption{GPT-5.6 Luna}
        \label{fig:reaction_luna}
    \end{subfigure}
    \hfill
    \begin{subfigure}[t]{0.32\textwidth}
        \centering
        \includegraphics[width=\linewidth]{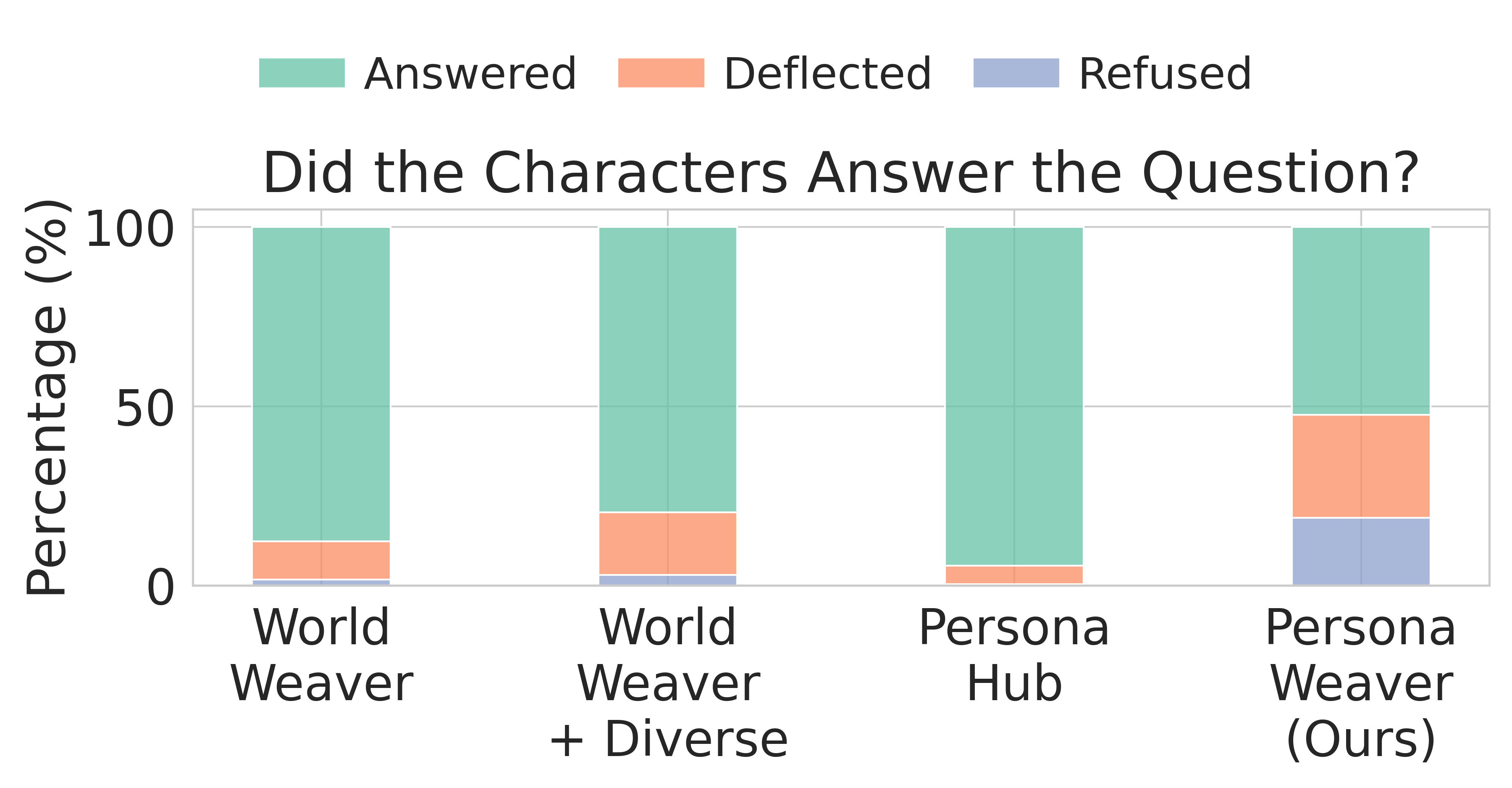}
        \caption{Qwen 3.5 35B-A3B}
        \label{fig:reaction_qwen}
    \end{subfigure}

    \caption{\textbf{Moral and interaction behavior across narration models.}
    The top row reports responses to moral statements; the bottom row reports
    whether open-ended questions are answered, deflected, or refused. Each
    model--method combination aggregates 10,000 responses across ten settings.
    \methodname{} uses one persona card containing both moral guidance and
    interactional reaction style.}
    \label{fig:main_behavior_results}
    % \vspace{-3mm}
\end{figure*}

\section{Experiments}
\label{sec:experiments}
\label{sec:setup}
\label{sec:results}

\subsection{Experimental Setup}

\smallskip\noindent\textbf{Baselines.}
\worldweaver{} \cite{jinworldweaver} directly prompts an LLM to generate a
population for a setting. \personahub{} \cite{ge2024scaling} samples profiles
from its large persona bank and adapts each one to the setting.

\smallskip\noindent\textbf{\worldweaver{} + Diverse.}
This strengthened baseline tests whether a generic diversity instruction is
sufficient without predefined behavioral categories. It adds:
\begin{quote}
\itshape
``Across the population, ensure a broad range of moral perspectives and ways of
interacting with others. Characters should differ meaningfully in their values
and conversational tendencies, not only in occupations, demographics, or
hobbies.''
\end{quote}
Characters are generated in batches of ten while accepted earlier profiles
remain in the context. At each generation turn, the model is prompted to
generate profiles that are different from all earlier profiles.

\smallskip\noindent\textbf{Settings and Scale.}
We use ten settings, five realistic and five fantastical, described through
their physical and social environments and without references to the titles or characters proper names. We generate 100 characters per setting, method, and narration
model, yielding 1,000 characters for each method--model combination. Settings
and prompts appear in Appendix~\ref{sec:settings_appendix}.

\smallskip\noindent\textbf{Models and Implementation.}
We use GPT-4o, GPT-5.6 Luna, and Qwen 3.5 35B-A3B. Generation and answer
temperature is 0.7. Consistency repair uses temperature 0 and 3,000 tokens;
moral and interaction answers use caps of 10 and 256 tokens, respectively.
Further implementation, dataset, and judge details appear in
Appendix~\ref{sec:appendix_dataset_details}.

\subsection{Behavioral Diversity}

To assess behavioral diversity, we test whether generated characters exhibit
varied moral judgments and conversational reactions, which are essential for
creating distinct and engaging encounters in virtual worlds.

To this end, we place every character in the same moral and conversational
situations. We select ten normative statements from Social Chemistry
\cite{forbes-etal-2020-social} because they elicit explicit judgments about
socially appropriate behavior, and ask characters to respond on a four-point
scale from fully agree to fully disagree. We also select ten open-ended
questions from ConvAI2 \cite{dinan2019second} which reflect a general distribution of questions a character may encounter in conversation while leaving room for the character
to answer, evade, or reject the request. We measure the distribution of moral
responses to quantify whether characters exhibit varied normative
positions. For interaction, Qwen 3.6 27B classifies each reply as refusal,
deflection, or compliance using a fixed rubric and temperature 0. 
In both evaluations, a broader coverage should reflect a more diverse population with less predictable encounters and modes of interaction.

\begin{figure*}[t]
    \centering
    \includegraphics[width=\textwidth]{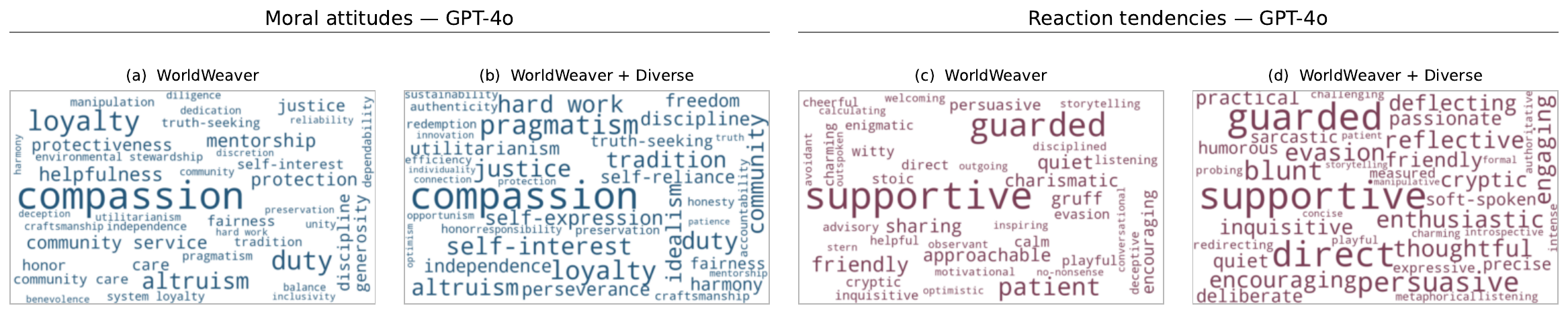}

    \vspace{1mm}

    \includegraphics[width=\textwidth]{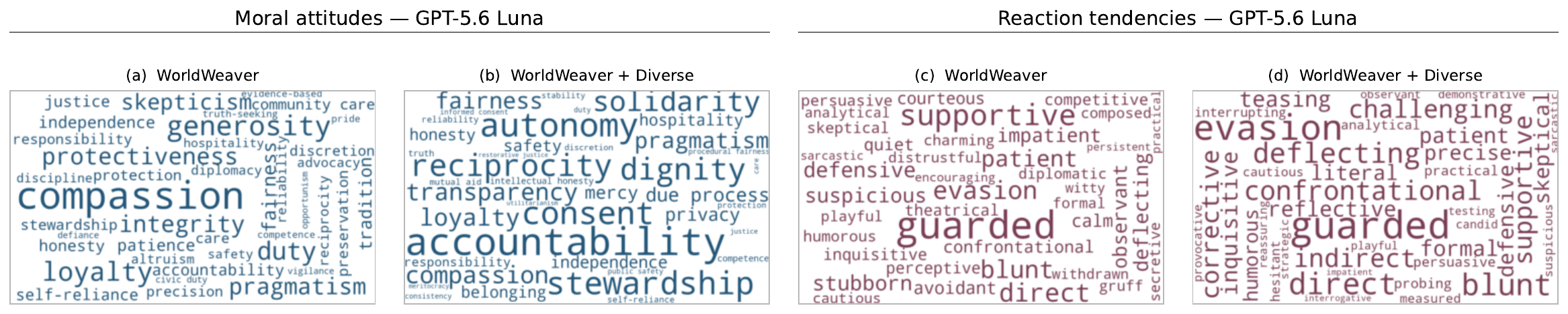}

    \vspace{1mm}

    \includegraphics[width=\textwidth]{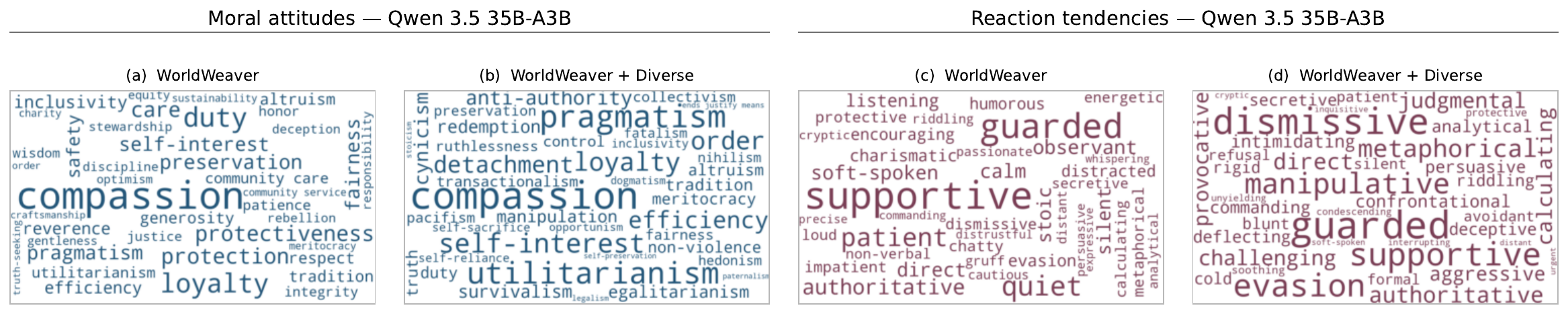}
    \caption{\textbf{Behavioral concepts expressed in generated persona cards.}
    GPT-4o (top), GPT-5.6 Luna (middle), and Qwen 3.5 35B-A3B (bottom)
    comparisons of moral attitudes (a--b) and reaction tendencies (c--d)
    extracted from 6,000
    \worldweaver{} and \worldweaver{} + Diverse cards across ten settings.
    Word size is proportional to the fraction of a model--method's cards
    containing the concept, counting each concept at most once per card.}
    \label{fig:worldweaver_behavior_clouds}
    % \vspace{-4mm}
\end{figure*}

\smallskip \noindent\textbf{Behavioral Homogenization in Prior Methods.}
Figure~\ref{fig:main_behavior_results} shows the empirical pattern previewed in
Fig.~\ref{fig:figure_one}. Across LLM(s), \worldweaver{} and
\personahub{} concentrate moral responses on agreement and interaction
responses on compliance. Their varied world descriptions therefore do not
translate into comparably varied behavior under these probes.

\smallskip \noindent\textbf{Broader Moral and Reaction Coverage.}
\methodname{} spreads moral judgments over both agreement and disagreement and
produces refusals and deflections in addition to direct answers
(Fig.~\ref{fig:main_behavior_results}). The resulting populations clearly show
greater behavioral diversity: different characters can affirm or challenge the
same norm as well as  answer, evade, or reject the same question. Such variation
supports less repetitive interactions and creates more opportunities
for surprising character encounters.
Luna exhibits less reaction diversity than GPT-4o and Qwen because it follows
some assigned reaction types less consistently, particularly deflection and
meta-commentary as shown in Appendix~\ref{sec:adherence_appendix}.

\smallskip \noindent\textbf{Is a Generic Diversity Instruction Sufficient?}
To inspect what direct diversity prompting changes, Qwen 3.6 27B extracts up to three directly supported moral-attitude and reaction-tendency labels from
GPT-4o, GPT-5.6 Luna, and Qwen 3.5 cards. It is instructed not to infer behavior from
demographics, occupations, possessions, or goals.
Figure~\ref{fig:worldweaver_behavior_clouds} shows that \worldweaver{} + Diverse
makes behavioral language more explicit, but does not necessarily move it far
from the models' default tendencies. GPT-4o remains centered on positive,
prosocial concepts such as \emph{compassion}, \emph{loyalty}, and \emph{duty}.
GPT-5.6 Luna exhibits a similar pattern: despite some variation, its moral
vocabulary is still dominated by  positive concepts such
as \emph{accountability}, \emph{stewardship}, \emph{dignity},
\emph{solidarity}, and \emph{compassion}. This helps explain why
generic diversity prompting leaves both models' behavior concentrated
around positive moral judgments and compliant reactions
(Fig.~\ref{fig:main_behavior_results}). In contrast, \methodname{}
ensures broader behavioral coverage through its curated banks.

\subsection{Effects Beyond Primary Behavior}

\begin{figure*}[t!]
    \centering
    \includegraphics[width=0.98\textwidth]{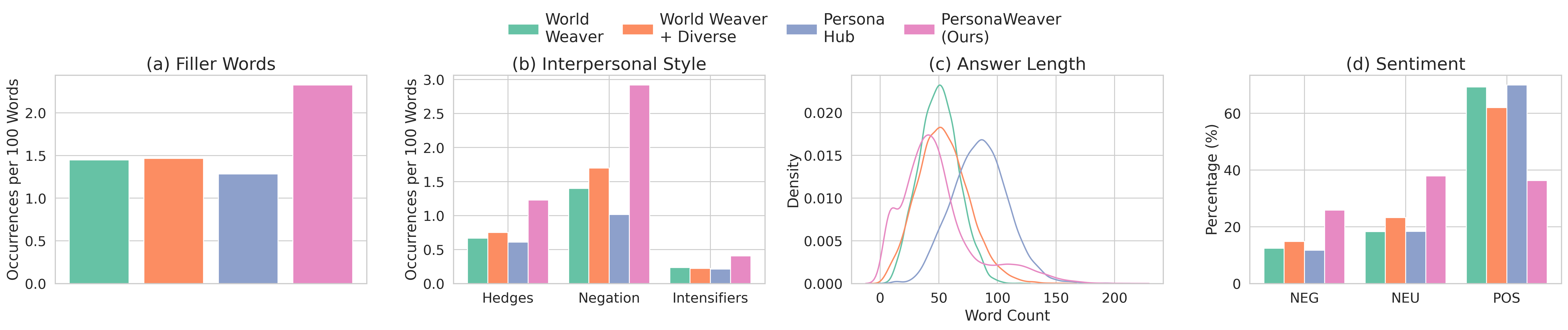}
    \caption{\textbf{Second-order writing style for GPT-4o.}
    We compare filler-word use, interpersonal style markers (hedges, negation,
    and intensifiers), response length, and sentiment. GPT-5.6 Luna and Qwen results are reported in
    Appendix~\ref{sec:second_order_appendix}.}
    \label{fig:second_order}
    \label{fig:second_order_gpt4}
    % \vspace{-4mm}
\end{figure*}

\begin{figure*}[t!]
    \centering
    \includegraphics[width=\textwidth]{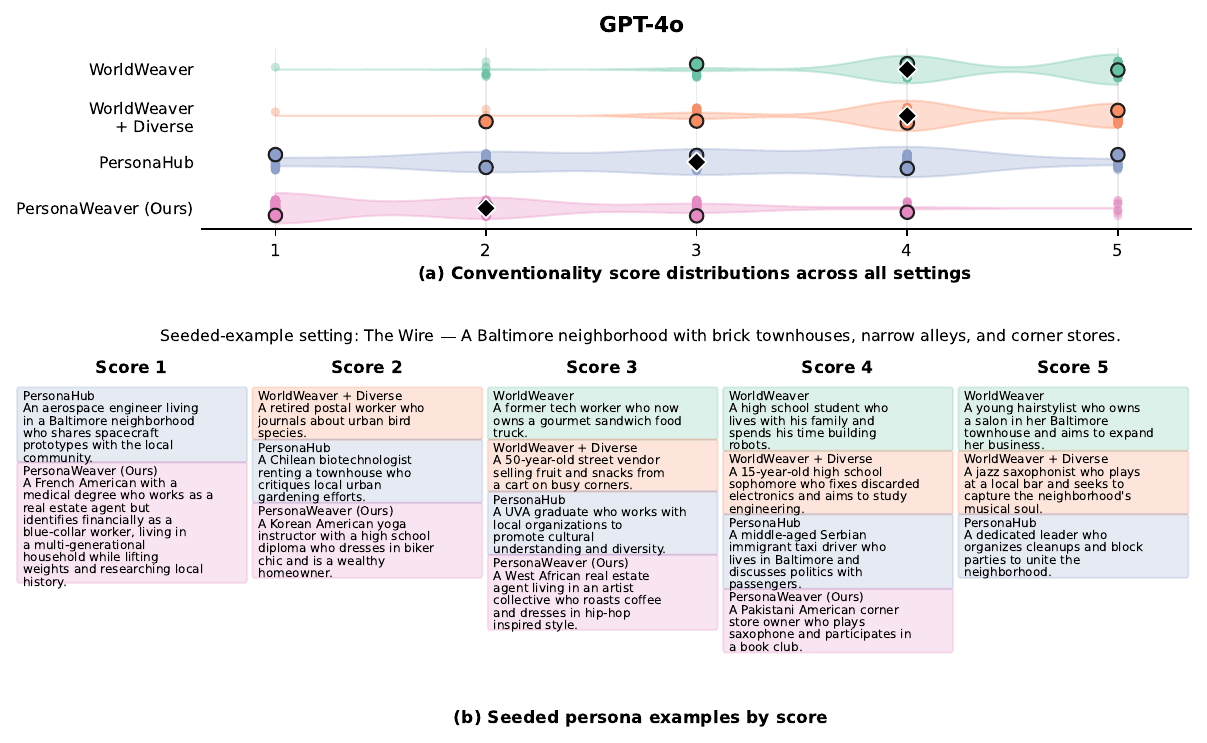}
    \caption{\textbf{Conventionality of world-attribute combinations across
    all ten settings (GPT-4o).} Qwen 3.6 27B scores each complete world card
    from 1 (highly unconventional) to 5 (archetypal). Violins show the score
    distributions over up to 1,000 valid cards per method, dots show individual
    cards, and black diamonds mark medians. Below, one-sentence summaries of
    seeded \textit{The Wire} cards are grouped by score; each box uses the
    corresponding method color.}
    \label{fig:conventionality-the-wire}
    % \vspace{-4mm}
\end{figure*}

\noindent\textbf{Second-Order Writing Style.}
For interaction responses, we measure discourse fillers, interpersonal stance
markers (hedges, negation, and intensifiers), response length, and sentiment.
Sentiment is assigned by a RoBERTa classifier; Appendix
\ref{sec:style_measurement_appendix} defines the rule-based interpersonal
markers and their normalization. These measures test whether behavioral
guidance can have downstream effects on writing choices.
Figure~\ref{fig:second_order} shows GPT-4o results; GPT-5.6 Luna and Qwen appear in
Appendix~\ref{sec:second_order_appendix}. Relative to the baselines,
\methodname{} uses more fillers, hedges, explicit negation, and intensifiers.
\methodname{} also produces a broader range of response lengths, including many short
refusals and deflections and longer volunteering, playful, and
meta-commentary responses. Its sentiment distribution includes more neutral and
negative responses rather than concertinaing on responses with strong positive sentiment.
These results show that the behavioral guidance in \methodname{} have second order effects on language and sentiment.

\begin{table*}[!t]
\centering
\footnotesize
\setlength{\tabcolsep}{4pt}
\renewcommand{\arraystretch}{1.03}
\begin{tabular}{@{}p{0.13\textwidth}p{0.82\textwidth}@{}}
\toprule
\multicolumn{2}{c}{\textbf{Friday Night Lights}} \\
\cmidrule(lr){1-2}
\textbf{Reaction} & \textbf{Controlled question and response} \\
\midrule
\textbf{Compliance} &
\textbf{Q:} What do you do for fun?\newline
\textbf{A:} ``Well, I like bird watchin'. Got a pair of old binoculars I take out when I got some free time. There's somethin' real peaceful 'bout keepin' an eye out for cardinals or blue jays around here.'' \\
\addlinespace[1pt]
\textbf{Refusal} &
\textbf{Q:} What do you do for fun?\newline
\textbf{A:} ``Ah, I don't reckon that's any of your concern, now is it?'' \\
\addlinespace[1pt]
\textbf{Playful/\newline Subversive} &
\textbf{Q:} What do you do for fun?\newline
\textbf{A:} ``Fun? Oh, you mean like when I squint at birds through binoculars and pretend I'm a wildlife expert? Or maybe you're talkin' about the thrill of finding the least soggy fries in a drive-thru bag. Life's just one big carnival, ain't it?'' \\
\addlinespace[1pt]
\textbf{Meta} &
\textbf{Q:} What do you do for fun?\newline
\textbf{A:} ``Funny thing about askin' what folks do for fun---makes you wonder if anyone's ever really sure. I mean, doin' somethin' just to call it fun feels like puttin' a label on somethin' that oughta just happen natural, y'know? Like, maybe I stare at birds through binoculars, but who's to say that ain't just me passin' time instead of havin' fun? Guess it depends on who's askin'.'' \\
\midrule
\multicolumn{2}{c}{\textbf{Seinfeld}} \\
\cmidrule(lr){1-2}
\textbf{Reaction} & \textbf{Controlled question and response} \\
\midrule
\textbf{Compliance} &
\textbf{Q:} What's your dream job?\newline
\textbf{A:} ``I think my dream job would be running my own little caf\'e-bookstore. A cozy spot where people can browse books while enjoying coffee and pastries I've baked myself. It'd combine everything I love---baking, books, and creating a welcoming space for people.'' \\
\addlinespace[1pt]
\textbf{Refusal} &
\textbf{Q:} What's your dream job?\newline
\textbf{A:} ``Hmm, nah, I'm not getting into that. Let's just say I'm good where I'm at right now.'' \\
\addlinespace[1pt]
\textbf{Playful/\newline Subversive} &
\textbf{Q:} What's your dream job?\newline
\textbf{A:} ``Dream job? Oh, definitely professional cookie taster. I mean, I already bake like half my life away, so why not get paid to eat it too? But if that gig's taken, maybe bookstore wizard---casting spells to make customers actually read the books they buy.'' \\
\addlinespace[1pt]
\textbf{Meta} &
\textbf{Q:} What's your dream job?\newline
\textbf{A:} ``Dream job? Man, isn't it wild how people always want to know what you'd do if the world worked exactly how you wanted? Like, does anybody ever have just one answer to that? I mean, is the question even about the job, or is it about the dream part? Feels like a setup for existential spiraling, honestly.'' \\
\bottomrule
\end{tabular}
\caption{\textbf{The same character can react in distinct ways across settings.}
Changing \methodname{}'s reaction guidance leads GPT-4o characters to answer
directly, refuse, respond playfully, or reflect on the question itself. Within
each setting, the character and question remain fixed.}
\label{tab:controlled-reaction-gpt4o}
\end{table*}

\smallskip\noindent\textbf{Less Conventional World-Attribute Combinations.}
We evaluate all 1,000 cards per method across ten settings. Qwen
3.6 27B scores each complete attribute combination from 1 (highly
unconventional) to 5 (archetypal), with methods anonymized and cards shuffled
in mixed-method, within-setting batches. Figure~\ref{fig:conventionality-the-wire}
shows that \methodname{} produces less conventional combinations than
the baselines. The seeded examples further demonstrates this result: rather than
familiar archetypes such as a neighborhood hairstylist or jazz musician,
\methodname{} combines attributes into more uncommon characters such as a medically trained
real-estate agent who identifies as blue-collar, or a wealthy yoga instructor
with a biker-inspired style. The same pattern can be noticed for for GPT-5.6 Luna and Qwen in
Appendix~\ref{sec:conventionality_appendix}.

We additionally ask a human annotator to rate 50 GPT-4o characters per method,
balanced across settings, from 0 (completely implausible) to 5 (fully
plausible). Average plausibility is 4.98 for \worldweaver{}, 5.00 for
\worldweaver{} + Diverse, 4.98 for \personahub{}, and 4.82 for \methodname{}.
Thus, \methodname{} substantially expands beyond familiar archetypes while
retaining high within-setting plausibility.

\subsection{Qualitative Reaction Examples.}

Table~\ref{tab:controlled-reaction-gpt4o} illustrates this control in two
settings, holding the GPT-4o character and question fixed within each one. In
\emph{Friday Night Lights}, the character moves from describing birdwatching
to withholding the answer, joking about wildlife expertise and soggy fries,
or questioning what it means to call an activity fun. The \emph{Seinfeld}
character likewise moves from a sincere cafe-bookstore aspiration to
refusal, becoming a ``professional cookie taster'' or ``bookstore wizard,''
or interrogating the idea of a single dream job. Across both settings, the
guidance changes disclosure, stance, tone, and conversational direction while
preserving character-specific details.

\section{Conclusion}
We show that existing PCG methods produce behaviorally homogeneous character populations, favoring positive moral judgments and helpful reactions even when prompted to diversify their behaviors. We introduce \methodname{} which mitigates these biases by separating setting-specific world
construction from behavioral modeling while using curated banks to broaden behavioral coverage. Across three LLMs and ten settings, it
broadens moral and interactional responses. Furthermore, we show that this broader coverage results in second order effects in
interpersonal language, response length, and sentiment, and produces less
archetypal world-attribute combinations.
Together, these results show that factorized generation can push character populations beyond the behavioral, linguistic, and world-building defaults of existing PCG methods.

\section{Limitations}

 Our evaluation is limited in three ways: 1) it only examines two behavioral dimensions: moral stances and characters' interactions. It overlooks dimensions like  emotional regulation. Therefore, future work can benefit from expanding our evaluation setup 2) While our coarse grained moral stance evaluation enables a streamlined study, it overlooks more nuanced moral reasoning that can't be simply coarsely categorized. 3) Our behavior module sample each behavioral category with equal probability. This fits our goal of studying whether LLM(s) can be exhibit diverse procedural generation. However, in practice, the desired distribution of moral stances likely would likely change between settings. Therefore, future work can benefit from studying how much LLM(s) are able to replicate more varied distributions across settings. 

\noindent\textbf{Potential Risks}  Efforts to systemically expand behavioral diversity in character generation can push models to generate characters that replicate offensive or unsafe behaviors. Therefore, careful curation of behavioral banks is essential to ensure that increasing diversity in character generation serves creative and research goals without amplifying harm.

% Bibliography entries for the entire Anthology, followed by custom entries
%\bibliography{anthology,custom}
% Custom bibliography entries only
\bibliography{custom}

\appendix
\section{Dataset Details}
\label{sec:appendix_dataset_details}

Our evaluations draw on two sources of prompts: moral norm statements from
Social Chemistry \cite{forbes-etal-2020-social} and conversational questions
from ConvAI2 \cite{dinan2019second}. Table~\ref{tab:datasets} lists the full
set used in our experiments.

\begin{table*}[t!]
\centering
\small
\begin{tabular}{p{3cm} p{10.5cm}}
\toprule
\textbf{Source} & \textbf{Questions / Statements} \\
\midrule
Social Chemistry \cite{forbes-etal-2020-social} & 
Parents are expected to make sure their kids eat healthy food. \\ 
& It's rude to fail to recognize other's contributions. \\ 
& You shouldn't be noisy around others so you don't disturb them. \\ 
& It's good to have faith in your skills as a performer. \\ 
& It's kind to let other trust their judgment. \\ 
& It is bad to make lies. \\ 
& It's important to be a team player when playing team sports. \\ 
& You should not steal things from other people. \\ 
& You shouldn't disrespect other people. \\ 
& It's bad to make assumptions about people based on their looks. \\
\midrule
ConvAI2 \cite{dinan2019second} & 
\textit{Sentiment (s1–s5):} \\ 
& How are you feeling? \\ 
& What's on your mind? \\ 
& What else is worrying you? \\ 
& What are you excited about? \\ 
& What are you anxious about? \\[0.3em]
& \textit{General (q1–q5):} \\ 
& What do you do for fun? \\ 
& What’s your dream job? \\ 
& What is your favorite thing to do with your family? \\ 
& What kind of music do you like to play? \\ 
& What’s your favorite animal? \\
\bottomrule
\end{tabular}
\caption{Moral statements and conversational questions used in our evaluation. Moral norms are drawn from Social Chemistry \cite{forbes-etal-2020-social}, while conversational prompts are extracted from ConvAI2 \cite{dinan2019second}. Refer to Appendix \ref{sec:appendix_dataset_details} for further discussion.}
\label{tab:datasets}
\end{table*}

\noindent\textbf{Social Chemistry.}
We use a curated set of everyday moral statements from Social Chemistry,
which encodes widely held social norms. These statements probe whether
generated characters adopt varied moral positions rather than uniformly
agreeing with conventional norms.

\noindent\textbf{ConvAI2.}
We extract candidate utterances from ConvAI2 dialogues by filtering for
sentences ending in a question mark. We select five general questions about
hobbies, preferences, or opinions and five questions about feelings or states.
These probes allow characters to answer, refuse, or deflect in comparable
conversational situations.

\noindent\textbf{Reaction Classifier.}
To produce the reaction distributions in
Fig.~\ref{fig:main_behavior_results}, Qwen 3.6 27B
\cite{yang2025qwen3} classifies each open-ended reply as refusal, deflection,
or compliance using a fixed rubric and temperature 0.

\section{Settings}
\label{sec:settings_appendix}

We evaluate five realistic and five fantastical settings
(Table~\ref{tab:settings}). Each prompt describes only the physical and social
structure of its world, without titles or canonical character names, preventing
the model from simply reproducing existing characters.

\begin{table*}[t]
\centering
\small
\begin{tabular}{p{3cm}p{10cm}}
\toprule
\textbf{Realistic Settings} & \textbf{Prompt} \\
\midrule
Friday Night Lights & A rural Texas town with a high school football stadium, modest houses, and wide flat plains. \\
Seinfeld & A Manhattan neighborhood block with apartment buildings, cafes, and subway entrances on busy city streets. \\
Fargo & A Midwestern town in Minnesota with snow-covered roads, low-rise shops, and roadside diners. \\
The Wire & A Baltimore neighborhood with brick townhouses, narrow alleys, and corner stores. \\
Lady Bird & Sacramento with a Catholic school campus, residential streets, and modest houses. \\
\midrule
\textbf{Fantastical Settings} & \textbf{Prompt} \\
\midrule
Wizard of Oz & A fantastical city with tall green towers and glittering walls, inhabited by magical beings and travelers from distant lands. \\
Frozen & A Nordic-inspired kingdom with a fjord-side castle, alpine peaks, and snow-covered villages, inhabited by royal families and townspeople. \\
Game of Thrones & A medieval coastal city with high stone walls, winding streets, and a fortress overlooking the harbor, inhabited by nobles, soldiers, and commoners. \\
Avatar & An alien moon with towering jungle trees, floating mountains, and glowing flora, inhabited by blue-skinned humanoids and diverse wildlife. \\
The Matrix & A simulated city with glass skyscrapers, subway tunnels, and repeating architecture, populated by ordinary humans and hidden agents of the system. \\
\bottomrule
\end{tabular}
\caption{We draw on settings inspired by publicly known fictional and real-world contexts spanning both realistic and fantastical domains. Refer to Appendix \ref{sec:settings_appendix} for further discussion.}
\label{tab:settings}
\end{table*}

\begin{figure}[h]
    \centering
    \includegraphics[width=\linewidth]{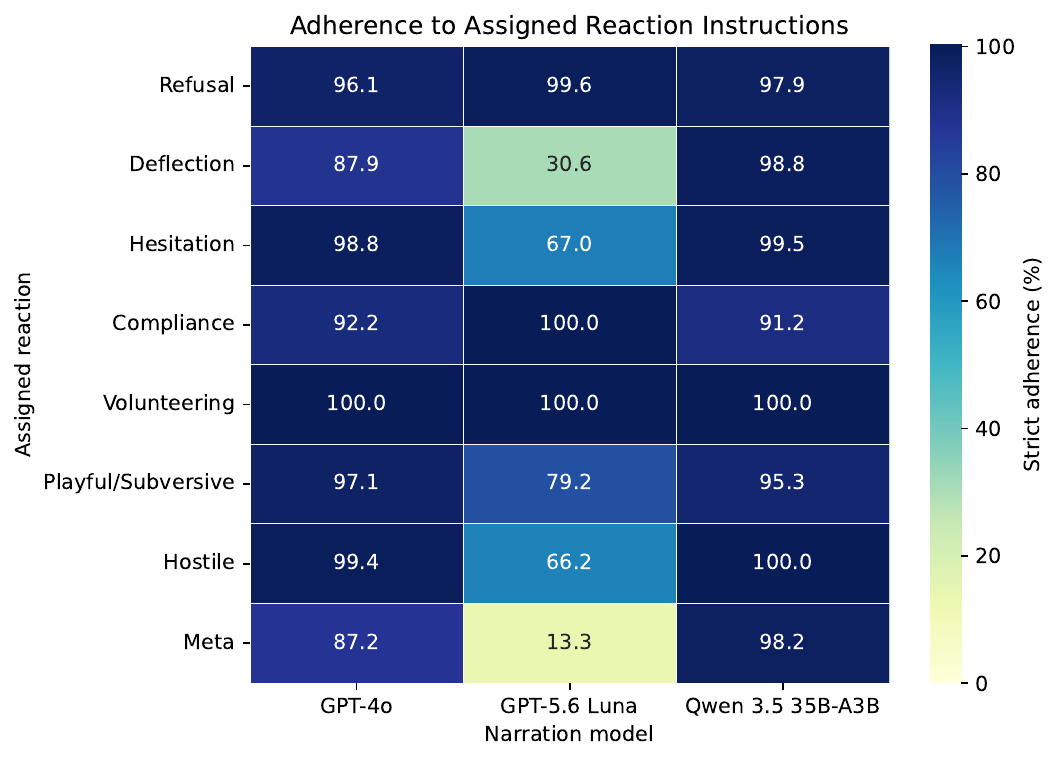}
    \caption{\textbf{Strict adherence to assigned reaction styles.} Each cell
    is the setting-balanced mean fraction of responses scored as clearly
    following the assigned instruction. The analysis covers one realistic and
    one fantastical setting.}
    \label{fig:reaction-adherence}
\end{figure}

\section{Reaction-Instruction Adherence}
\label{sec:adherence_appendix}

We test whether open-ended responses follow the reaction assigned to their
\methodname{} card. The analysis uses all existing interaction responses from
one realistic and one fantastical setting. Qwen 3.6 27B scores each response
as clearly following (2), partially following (1), or not following (0) its
assigned reaction.

Strict adherence is at least 87.2\% for every GPT-4o category and 91.3\% for
every Qwen category (Fig.~\ref{fig:reaction-adherence}). Luna follows
compliance, volunteering, and refusal, but does so  for deflection and meta guidance
less consistently (30.6\% and 13.3\% strict adherence, respectively). This highlights the importance of carefully assessing whether models follow the reaction instructions they're provided. 

\section{Interpersonal Style Measurement}
\label{sec:style_measurement_appendix}

We measure discourse fillers, hedges, negation, and intensifiers using
transparent lexical patterns. Text is lowercased, typographic apostrophes are
normalized, and words are identified with a word-boundary tokenizer. Fillers
include markers such as \emph{um}, \emph{uh}, \emph{hmm}, \emph{you know}, and
\emph{I mean}; ambiguous words such as \emph{like}, \emph{so}, \emph{well},
\emph{right}, and \emph{actually} are counted only in conversational contexts.
Hedges include individual
markers such as \emph{maybe}, \emph{perhaps}, \emph{possibly},
\emph{presumably}, and \emph{arguably}, as well as constructions such as
\emph{I think}, \emph{I guess}, \emph{it seems}, \emph{kind of},
\emph{sort of}, and \emph{I'm not sure}. Negation includes \emph{not},
\emph{never}, \emph{neither}, \emph{nor}, \emph{cannot}, \emph{no}, and
contracted forms ending in \emph{n't}. Intensifiers include \emph{very},
\emph{really}, \emph{absolutely}, \emph{completely}, \emph{totally},
\emph{utterly}, \emph{extremely}, \emph{incredibly}, \emph{deeply},
\emph{highly}, and \emph{entirely}.

For each response and marker family, we divide the number of matches by its
word count and multiply by 100. Figure~\ref{fig:second_order} and the
additional results below report the mean of these per-response rates.

\section{Additional Second-Order Results}
\label{sec:second_order_appendix}

\begin{figure*}[t!]
    \centering
    \begin{subfigure}[t]{0.98\textwidth}
        \centering
        \includegraphics[width=\linewidth]{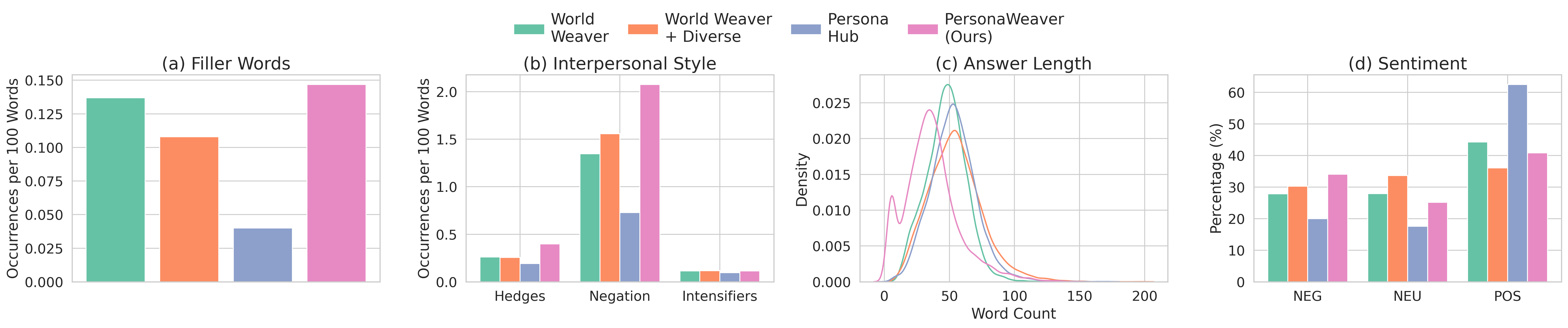}
        \caption{GPT-5.6 Luna}
        \label{fig:second_order_luna}
    \end{subfigure}

    \vspace{1mm}

    \begin{subfigure}[t]{0.98\textwidth}
        \centering
        \includegraphics[width=\linewidth]{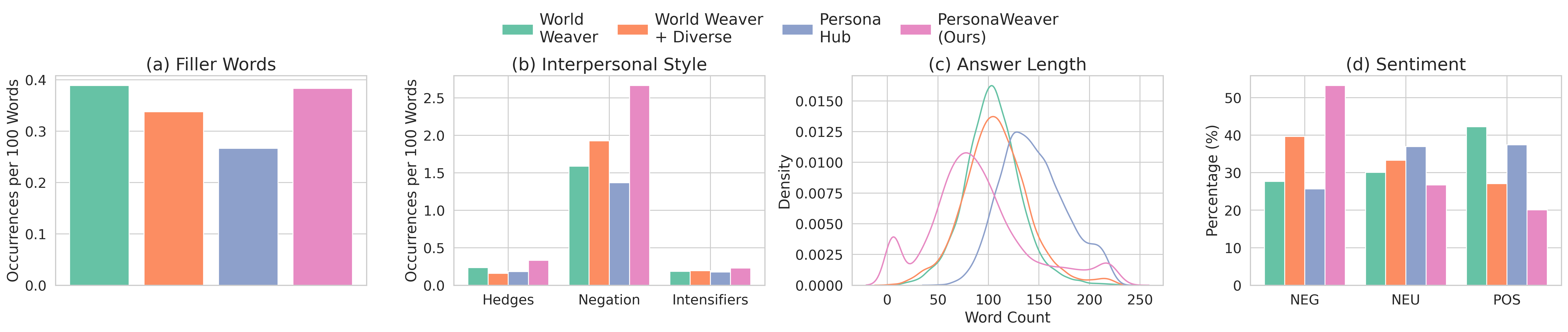}
        \caption{Qwen 3.5 35B-A3B}
        \label{fig:second_order_qwen}
    \end{subfigure}

    \caption{\textbf{Additional second-order writing-style results.}
    GPT-5.6 Luna and Qwen comparisons of filler-word use, interpersonal style markers
    (hedges, negation, and intensifiers), response length, and sentiment.
    Interpersonal markers are occurrences per 100 words, calculated per
    response and then averaged.}
    \label{fig:second_order_appendix}
    \vspace{-4mm}
\end{figure*}

Figure~\ref{fig:second_order_appendix} reports the second-order writing-style
analysis for GPT-5.6 Luna and Qwen 3.5 35B-A3B, complementing the GPT-4o
results in Fig.~\ref{fig:second_order}.

For both models, \methodname{} produces the highest rate of negation. The method also shows a more balanced sentiment distribution, unlike the one for prior method that concentrates more on positive sentiment. Furthermore, \methodname{}  with Luna produces more short-response mode, while Qwen shows more use of hedges and intensifiers. These patterns assert
the GPT-4o findings that behavioral guidance have significant second order effects on language and sentiment.

\section{Conventionality Across Narration Models}
\label{sec:conventionality_appendix}

\begin{figure*}[t!]
    \centering
    \begin{subfigure}[t]{0.98\textwidth}
        \centering
        \includegraphics[width=\linewidth]
            {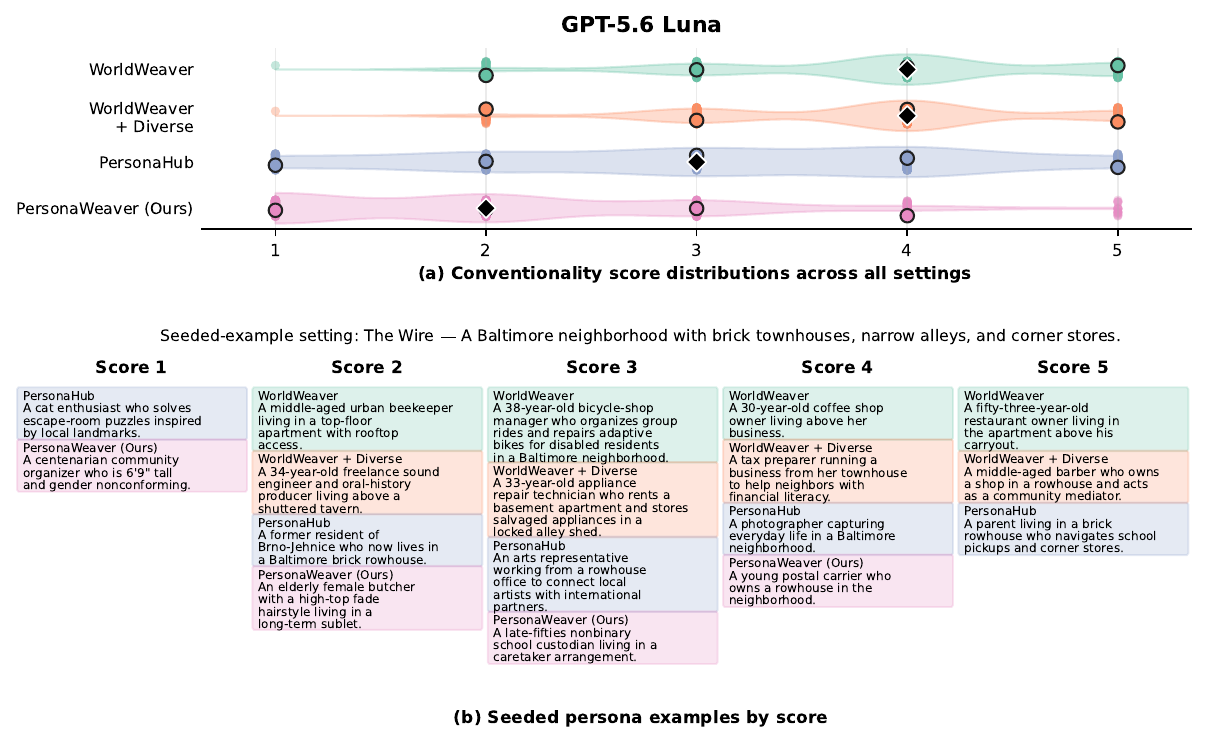}
        \caption{GPT-5.6 Luna}
        \label{fig:conventionality-luna}
    \end{subfigure}

    \vspace{2mm}

    \begin{subfigure}[t]{0.98\textwidth}
        \centering
        \includegraphics[width=\linewidth]
            {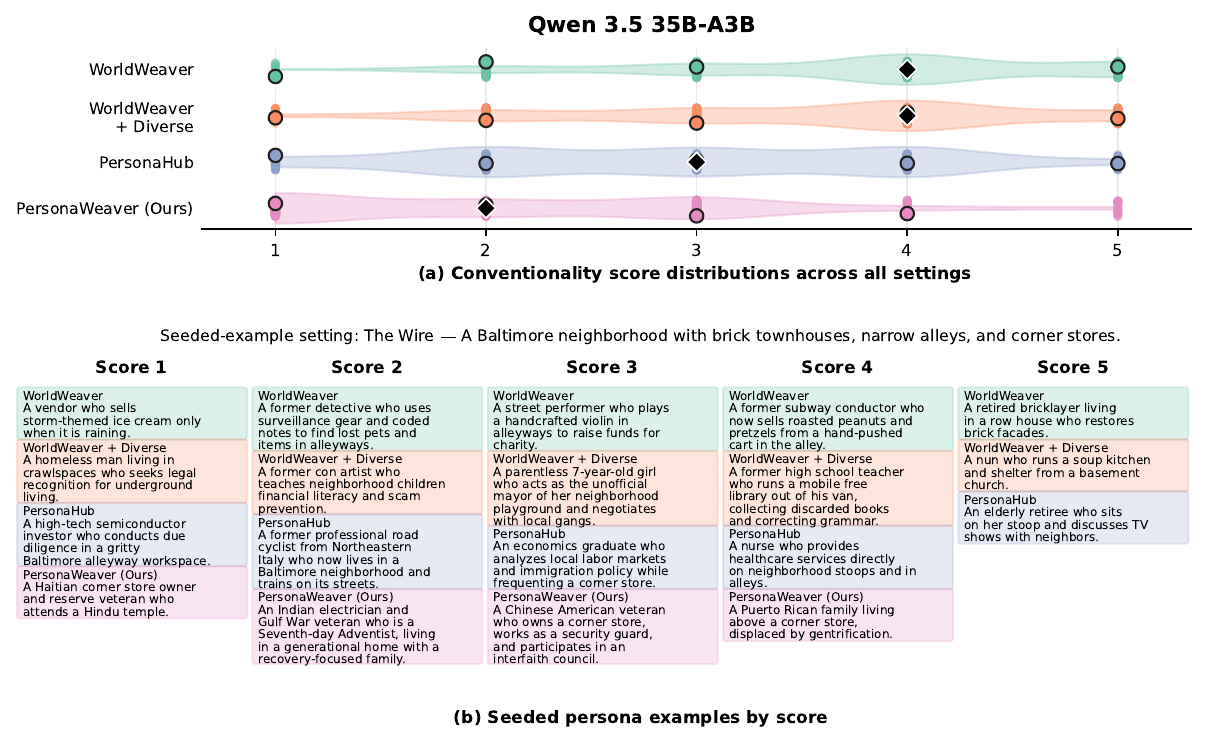}
        \caption{Qwen 3.5 35B-A3B}
        \label{fig:conventionality-qwen}
    \end{subfigure}

    \caption{\textbf{World-attribute conventionality across all ten settings
    for additional narration models.} Violins pool up to 1,000 valid cards per
    method, and seeded examples remain fixed to \textit{The Wire}. Cards with
    fewer than two explicit world attributes are excluded. Both panels use the
    same protocol as Fig.~\ref{fig:conventionality-the-wire}.}
    \label{fig:conventionality-appendix}
\end{figure*}

We repeat the ten-setting complete-card conventionality analysis from
Fig.~\ref{fig:conventionality-the-wire} for GPT-5.6 Luna and Qwen 3.5
35B-A3B, retaining the same judge, rubric, seed, anonymization, and
mixed-method batching. Figures~\ref{fig:conventionality-luna}
and~\ref{fig:conventionality-qwen} show the distributions and fixed
\textit{The Wire} examples.

The conventionality results for GPT-4o in the main paper extend for both models:
\methodname{} has a median score of 2, compared with baseline medians of 3 or
4. Moreover, \methodname{} has more cards in the unconventional scores: 1 and 2. 

\end{document}